\documentclass[letterpaper,10pt,conference]{IEEEtran}

\usepackage[english]{babel}
\usepackage{amssymb}
\usepackage{amsmath}
\usepackage{hyperref}
\usepackage{cleveref}
\usepackage{graphicx}
\usepackage{latexsym}
\usepackage{algorithm}
\usepackage{algpseudocode}
\usepackage{todonotes}
\usepackage{newclude}
\usepackage{multicol}
\usepackage{tabularx}
\usepackage{booktabs}
\usepackage{threeparttable}

\usepackage{enumitem}
\usepackage{listings}
\usepackage{combelow}
\usepackage{subfigure}
\usepackage{bigfoot}
\usepackage{newtxmath}

\usepackage{multirow}
\newcolumntype{C}[1]{>{\centering}m{#1}}

\usepackage[acronym]{glossaries}
\newacronym{gps}{GPS}{Global Positioning System}
\newacronym{imu}{IMU}{Inertial Measurement Unit}
\newacronym{gnss}{GNSS}{Global Navigation Satellite System}
\newacronym{rtk}{RTK}{Real-time Kinematic}
\newacronym{sdk}{SDK}{Software Development Kit}
\newacronym{vo}{VO}{Visual Odometry}
\newacronym{ins}{INS}{Inertial Navigation System}
\newacronym{lidar}{LiDAR}{Light Detection and Ranging}
\newacronym{fmcw}{FMCW}{Frequency-Modulated Continuous-Wave}

\crefname{table}{Table}{Tables}
\crefname{figure}{Figure}{Figures}
\crefname{section}{Section}{Sections}

\usepackage{url}
\newcommand\rurl[1]{%
  \href{http://#1}{\nolinkurl{#1}}%
}

\usepackage[binary-units]{siunitx}
\IEEEoverridecommandlockouts

\title{Real-time Kinematic Ground Truth \\for the Oxford RobotCar Dataset}
\author{Will Maddern, Geoffrey Pascoe, Matthew Gadd, Dan Barnes, Brian Yeomans, and Paul Newman
\\
\\
Oxford Robotics Institute, Dept. Engineering Science, University of Oxford, UK
\\
\texttt{robotcardataset@robots.ox.ac.uk}
}

\begin{document}
\maketitle

\begin{abstract}
We describe the release of reference data towards a challenging long-term localisation and mapping benchmark based on the large-scale \textit{Oxford RobotCar Dataset}.
The release includes \num{72} traversals of a route through Oxford, UK, gathered in all illumination, weather and traffic conditions, and is representative of the conditions an autonomous vehicle would be expected to operate reliably in.
Using post-processed raw \acrshort{gps}, \acrshort{imu}, and static \acrshort{gnss} base station recordings, we have produced a globally-consistent centimetre-accurate ground truth for the entire year-long duration of the dataset.
Coupled with a planned online benchmarking service, we hope to enable quantitative evaluation and comparison of different localisation and mapping approaches focusing on long-term autonomy for road vehicles in urban environments challenged by changing weather.
\end{abstract}

\section{Introduction}
\label{sec:introduction}

For real-world autonomous driving systems, the challenges of reliable localisation and mapping in changing conditions using vision and \gls{lidar} are well documented, and many impressive solutions have been proposed.
However, most of these approaches are evaluated on small-scale datasets with only a few examples of challenging conditions, or medium-scale datasets with limited variation in driving conditions.
The well-known KITTI dataset and associated benchmarks~\cite{geiger2012we} uses data gathered in good conditions over a period of five days, and hence only represents a small fraction of the conditions an autonomous vehicle can expect to encounter over its operational lifetime.

In this paper we present an important prerequisite in the form of the underlying reference data for the localisation benchmark for autonomous vehicles which we are developing using the \textit{Oxford RobotCar Dataset}~\cite{RobotcarDatasetIJRR}.
This large-scale dataset consists of image, \gls{lidar}, and \gls{gps} data collected over a year of driving a repeated route in Oxford, UK, covering over \SI{1000}{\kilo\metre} of total distance.
A wide range of variation including illumination, weather, dynamic objects, seasonal changes, roadworks, and building construction were captured during the course of data collection.
To build a localisation ground truth, we have post-processed the raw \gls{gps} and \gls{imu} data with \gls{gnss} base station recordings to produce a centimetre-accurate \gls{rtk} solution.
We offer the corrected \gls{rtk} solutions for a subset of traversals, and withhold the remaining ground truth with a view towards an online benchmarking service similar to the KITTI Vision Benchmark suite~\cite{geiger2012we}.
By providing this important prerequisite for such a benchmark where researchers can quantitatively evaluate and compare localisation and mapping approaches on a challenging large-scale dataset, we hope to accelerate development of long-term autonomy for future autonomous vehicles.

\begin{figure*}
  \centering
  \includegraphics[width=\textwidth]{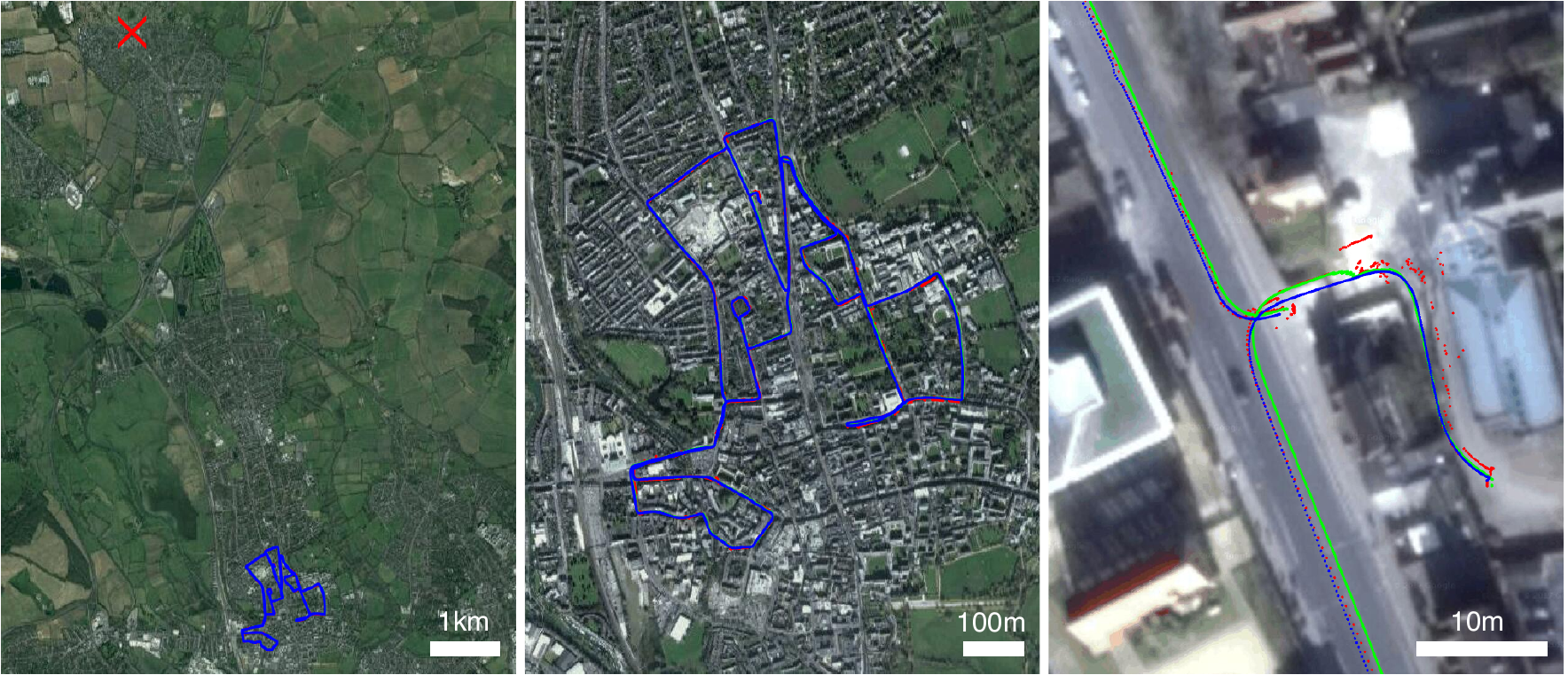}
  \caption{RTK ground truth for a single traversal from the \textit{Oxford RobotCar Dataset}.
  Left: The GNSS base station location, shown with the red X, relative to the data collection route.
  Center: \num{80} traversals (\num{72} released and \num{7} withheld) of the \SI{10}{\kilo\metre} route have been processed for localisation ground truth.
  Right: Quality of the \gls{rtk} position solution (blue) compared to the \gls{gps}-only (red) and \gls{gps}-inertial (green).
  A subset of seven \gls{rtk} ground truth files have been withheld to form a planned localisation benchmark which is under development.}
  \label{fig:gps-summary}
\end{figure*}

\section{Related Benchmarks}
\label{sec:related}

A number of urban driving datasets have been made available~\cite{pandey2011ford, pfeiffer2013exploiting, blanco2014malaga,waymo_open_dataset,yan2019eu,barnes2019oxford}, including datasets that focus on specific challenging scenarios~\cite{meister2012outdoor} and general apperance change over time~\cite{Badino_ICRA12}, but none of these offer a benchmarking service for comparison of results.

The KITTI dataset~\cite{geiger2013vision} offers a comprehensive benchmarking suite for stereo, optical flow, visual odometry, object detection and tracking.
However, KITTI data was collected only over a period of \num{5} days and does not contain challenging weather conditions, nor does it revisit the same location at different times for evaluating localisation.
Similarly, the Cityscapes dataset and benchmark~\cite{cordts2016cityscapes} contains stereo imagery from a wide range of locations but does not revisit locations at different times.

The most similar benchmark is the VPRiCE Challenge\footnote{\url{http://roboticvision.atlassian.net/wiki/spaces/PUB/pages/14188617}} which contains several different sequences to evaluate loop closure
in challenging conditions, including matching across night to day and between seasons.
However, the localisation metric for evaluation is precision vs recall, which does not incorporate the true 6DoF metric pose relative to the prior map, and the locations are only traversed twice. 
In contrast, the reference data presented in this paper evaluates full 6DoF pose estimation over \num{72} traversals of the route over the period of a year, totalling approximately \SI{650}{\kilo\metre} of driving.

\section{The Oxford RobotCar Dataset(s)}
\label{sec:rcd}

Our reference data and benchmark builds upon the \textit{Oxford RobotCar Dataset}~\cite{RobotcarDatasetIJRR}, one of the largest available datasets for autonomous driving research. 
It consists of over \SI{20}{\tera\byte} of vehicle-mounted monocular and stereo imagery, 2D and 3D \gls{lidar}, and inertial and \gls{gps} data collected over a year of driving in Oxford, UK.
More than 100 traversals of a \SI{10}{\kilo\metre} route illustrated in~\cref{fig:gps-summary} were performed over this period to capture scene variation over a range of timescales, from the \SI{24}{\hour} day/night illumination cycle to long-term seasonal variations.
For more details we refer the reader to~\cite{RobotcarDatasetIJRR}.

We also refer interested readers to the \textit{Oxford Radar RobotCar Dataset}~\cite{barnes2019oxford}.
While focused on millimetre-wave \gls{fmcw} scanning radar, this dataset provides over \SI{280}{\kilo\metre} of new sensor data common to the original \textit{Oxford RobotCar Dataset} as well as additional 3D \gls{lidar} data

\section{RTK Ground Truth}
\label{sec:RTK}

We have produced the localisation ground truth using low-level raw \gls{gps} and IMU data collected by the NovAtel SPAN-CPT mounted to the RobotCar.
The SPAN-CPT is a high-accuracy inertial navigation system (INS) equipped with dual \gls{gps} antennas, fibre-optic gyroscopes and MEMS accelerometers\footnote{\rurl{https://www.novatel.com/products/span-gnss-inertial-systems/span-combined-systems/span-cpt}}.
The raw recordings were not released as part of the original dataset.

GNSS base station data was obtained from the UK Ordnance Survey\footnote{\url{http://www.ordnancesurvey.co.uk/gps/os-net-rinex-data/}}, consisting of RINEX corrections updated at \SI{1}{\hertz}.
The recordings were sourced from the static base station in Kidlington, UK, approximately \SI{8.2}{\kilo\metre} from central Oxford.
Fig.~\ref{fig:gps-summary} illustrates the location of the base station relative to the data collection area.
Crucially, the base station remained stationary for the entire duration of the data collection and hence provides a consistent position reference for \gls{rtk} corrections.

We have post-processed the raw GPS, IMU and GNSS base station data using NovAtel Inertial Explorer\footnote{\url{novatel.com/products/software/inertial-explorer/}} to form an optimised corrected \gls{rtk} solution for all trajectories using tightly-coupled GNSS and IMU observations.
The \gls{rtk} solution provides corrected global position and orientation at \SI{10}{\hertz}.
Fig.~\ref{fig:gps-summary} illustrates the quality of the \gls{rtk} corrected solution in comparison to the \gls{gps}-only and \gls{gps}-inertial solutions available at runtime.
The estimated positioning errors of the \gls{rtk} solution are typically less than \SI{15}{\centi\metre} in latitude and longitude and less than \SI{25}{\centi\metre} in altitude, and the orientation errors are less than \SI{0.01}{\degree} in pitch and roll and \SI{0.1}{\degree} in yaw.

\begin{table*}[]
\centering
\begin{tabular}{|c|cccc|cccc|}
\hline
\multirow{2}{*}{\textbf{Method}} & \multicolumn{4}{c|}{\textbf{Position Error (m)}} & \multicolumn{4}{c|}{\textbf{Orientation Error (deg)}} \\
 & Lat & Lon & Alt & \textbf{Total} & Roll & Pitch & Yaw & \textbf{Total}\\ \hline
GPS & 2.88 & 1.78 & 8.02 & \textbf{8.71} & - & - & - & \textbf{-} \\ \hline
GPS+Inertial & 1.24 & 1.06 & 6.92 & \textbf{7.11} & 3.06 & 0.25 & 2.45 & \textbf{3.93} \\ \hline
\end{tabular}
\caption{Example \gls{gps} error evaluation relative to \gls{rtk} ground truth for a single route.}
\label{tab:errors}
\end{table*}

\section{Error Evaluation}
\label{sec:errors}

To benchmark localisation performance we plan to evaluate two metrics: root mean square (RMS) position and orientation errors, and uncertainty estimation.
The RMS errors are evaluated as follows:

\begin{equation}\label{eqn:rms-error}
  \sqrt{\frac{1}{N}\sum_{k=1}^{N}\left(\mathbf{\hat{x}}_{k}-\mathbf{x}_{k}\right)^{\mathrm{T}}\left(\mathbf{\hat{x}}_{k}-\mathbf{x}_{k}\right)}
\end{equation}

\noindent
where $\mathbf{\hat{x}}_{k}\in\mathbb{R}^{3\times1}$ is the position or orientation estimate at time $k$ and $\mathbf{x}_{k}$ is the ground truth position or orientation (from the \gls{rtk} solution), for a total of $N$ estimates.

\cref{tab:errors} presents example results for the errors computed with \gls{gps}-only and \gls{gps}-inertial solutions relative to the \gls{rtk} ground truth on one trajectory.

\section{Archives Amended}
\label{sec:archives}

In total, we are releasing \gls{rtk} solutions for \num{72} forays.
We exclude from release seven \textit{RobotCar seasons} runs for which careful ground truth pose is curated in~\cite{sattler2018benchmarking}, namely:
\begin{itemize}
    \item \verb|2014-12-16-09-14-09| -- dawn
    \item \verb|2015-02-20-16-34-06| -- dusk
    \item \verb|2014-12-10-18-10-50| -- night
    \item \verb|2014-12-17-18-18-43| -- night+rain
    \item \verb|2015-05-22-11-14-30| -- overcast (summer)
    \item \verb|2015-11-13-10-28-08| -- overcast (winter)
    \item \verb|2015-02-03-08-45-10| -- snow
    \item \verb|2015-03-10-14-18-10| -- sun
\end{itemize}
Where solutions for the eighth \verb|25 Nov 2014 - rain| run were not available.
This decision was made in order to ensure that there is a third party benchmark which calculates performance against a hidden ground truth signal for a swathe of challenging conditions from the original \textit{Oxford RobotCar dataset}.

\section{Software Development Kit}
\label{sec:sdk}

The original \gls{sdk}\footnote{\rurl{https://github.com/ori-mrg/robotcar-dataset-sdk}} has been updated to allow existing methods to use the \gls{rtk} solutions where \gls{vo} or \gls{ins} poses were already used interchangeably.
Specifically, the building of pointclouds and projection of laser scans into cameras images can now be performed with the \gls{rtk} solutions using the corresponding Python scripts:
\begin{itemize}
    \item \verb|build_pointcloud.py| and 
    \item \verb|project_laser_into_camera.py|
\end{itemize}
as well as MATLAB functions:
\begin{itemize}
    \item \verb|BuildPointcloud.m| and 
    \item \verb|ProjectLaserIntoCamera.m|.
\end{itemize}
The reader is referred to the dataset documentation\footnote{\url{https://robotcar-dataset.robots.ox.ac.uk/documentation}} and the \gls{sdk} documentation for more information.

\section{Conclusions}
\label{sec:conclusions}

We have presented the prerequisite reference data towards the planned \textit{Oxford RobotCar Long-Term Autonomy Benchmark}, a new dataset for evaluating long-term localisation and mapping approaches for autonomous vehicles in dynamic urban environments.
We expect to offer the benchmark as part of the \textit{Oxford RobotCar Dataset} website\footnote{\url{http://robotcar-dataset.robots.ox.ac.uk}} in the near future.

\bibliographystyle{IEEEtran}
\bibliography{biblio}

\end{document}